# Predicting property damage from tornadoes with zero-inflated neural networks

Jeremy Diaz*, Maxwell B. Joseph

*Earth Lab, Cooperative Institute for Research in Environmental Sciences, University of Colorado Boulder, United States*

ABSTRACT

Tornadoes are the most violent of all atmospheric storms. In a typical year, the United States experiences hundreds of tornadoes with associated damages on the order of one billion dollars. Community preparation and resilience would benefit from accurate predictions of these economic losses, particularly as populations in tornado-prone areas increase in density and extent. Here, we use a zero-inflated modeling approach and artificial neural networks to predict tornado-induced property damage using publicly available data. We developed a neural network that predicts whether a tornado will cause property damage (out-of-sample accuracy = 0.821 and area under the receiver operating characteristic curve, AUROC, = 0.872). Conditional on a tornado causing damage, another neural network predicts the amount of damage (out-of-sample mean squared error = 0.0918 and $R^2$ = 0.432). When used together, these two models function as a zero-inflated log-normal regression with hidden layers. From the best-performing models, we provide static and interactive gridded maps of monthly predicted probabilities of damage and property damages for the year 2019. Two primary weaknesses include (1) model fitting requires log-scale data which leads to large natural-scale residuals and (2) beginning tornado coordinates were utilized rather than tornado paths. Ultimately, this is the first known study to directly model tornado-induced property damages, and all data, code, and tools are publicly available. The predictive capacity of this model along with an interactive interface may provide an opportunity for science-informed tornado disaster planning.

## 1. Introduction

The United States experiences more tornadoes every year than any other country in the world, with the annual average of cumulative tornado-induced property damage at nearly one billion US dollars (Changnon, 2009). However, the distribution of property damages is heavy-tailed: the annual average has been exceeded by costs from single severe tornadoes, such as the 2011 Joplin tornado, which caused $2.8 billion dollars in damages (Storm Prediction Center, 1950). The damages resulting from tornadoes is a function of the physical properties of storms and societal factors such as population density, property values, and quality of building materials (Kunkel et al., 1999; Changnon et al., 2000; American Meteorological Society). Independent of physical changes in the number, distribution, or intensity of tornadoes, increasing property values, population density, and manufactured home density may have contributed to increases in tornado damages in recent decades (Kunkel et al., 1999; Changnon et al., 2000; American Meteorological Society; Kellner and Dev, 2014; Ashley and Strader, 2016; Ashley et al., 2014). These societal factors may be useful for predicting future tornado damages under different scenarios, with applications to development planning (Godschalk, 2003), natural disaster asset prepositioning (Salmerón and Apte, 2010), refinement of public warning systems (Stensrud et al., 2009), the property-casualty insurance industry (Changnon, 2009), and disaster response coordination (McEntire, 2002).

Here, we model tornado-induced property damages from the NOAA Storm Events Database as a function of 50 known and hypothesized drivers of damages, compiled from publicly available data. While previous research has identified variables that may be important (Kunkel et al., 1999; Changnon et al., 2000; American Meteorological Society; Kellner and Dev, 2014; Ashley and Strader, 2016; Ashley et al., 2014), the assumption of linearity in the effects is limiting (Kahn, 2005). The functional relationships among these proposed drivers and tornado damages are uncertain and possibly nonlinear, requiring a flexible means of predicting future observations as an unknown function of possibly interacting inputs. We use artificial neural networks to predict tornado-induced property damage over 22 of the most recent years (1997–2018) as a function of explanatory variables identified in the tornado risk literature.

* Corresponding author.
*E-mail address:* jeremy.diaz@colorado.edu (J. Diaz).

The universal function approximation capacity of artificial neural networks (Hornik et al., 1989, 1991) provides a means to learn the possibly complex functional relationships between proposed drivers and property damage. Previously, neural networks have been used to predict the occurrence of tornadoes from radar data (Marzban and Stumpf, 1996; Marzban, 2000), the occurrence of damaging winds (Marzban and Stumpf, 1998), and the trajectories of hurricanes (Giffard-Roisin et al., 2018). Conditional on the occurrence of a natural disaster, neural networks have proved useful for assessing damages, for instance after the December 2004 tsunami in Aceh, Sumatra (Aitkenhead et al., 2007). The application of machine learning to the prediction of potential tornado damages appears promising, though nascent in the current literature despite their increasing popularity and repeated calls for state of the art methods in the field of natural hazards (Van de Walle and Turoff, 2007; Pelletieret al., 2015) - with very recent calls for deep learning in the field of earth science (Reichstein et al., 2019).

The NOAA Storm Events Database provides an opportunity to train a model with historical tornado-induced property damages with the goal of generating future predictions that are useful for disaster relief and development scenario-analysis. Additionally, the Multi-Resolution Land Cover Characteristics (MRLC) Consortium's National Land Cover Database and U.S. Census Bureau's American Community Survey provide the ability to augment raw weather records with information on the physical and social status of affected areas.

Along with developing a neural network approach for predicting outcomes in a long-tailed, non-linear, and zero-inflated case, we compare the usefulness of several variable sets and produce 2019 predictions of damage occurrence and property losses conditioned on a tornado occurring. These predictions are presented in the form of maps, with additional emphasis on well-studied states, such as Kansas (Strader and Ashley, 2018), Alabama (Strader and Ashley, 2018; Ernest Agee and Taylor, 2019; Strader et al., 2016), Illinois (Ashley et al., 2014; Hall and Ashley, 2008; Strader et al., 2016; Rosencrants and Ashley, 2015), Oklahoma (Strader et al., 2016; Rosencrants and Ashley, 2015; Ernest Agee and Taylor, 2019), and Florida (Elsner et al., 2018; Hagemeyer and Schmocker, 1991). In addition to in-print visualizations, we present prototype dashboards that communicate these predictions accessibly (online, open source) and in a way that could be used by planners without expertise in machine learning or spatial analysis (Andre and Smith, 2003).

## 2. Methods

### *2.1. Data acquisition and fusion*

#### *2.1.1. Tornado data*

To develop the predictive models, we acquired data from past tornado events from NOAA's Storm Events Database (hereafter "Storm Events") (National Centers for Environmental Information, 1997), which includes information about where and when a tornado occurred, tornado path length and width, and how much property damage it caused. Property damage is recorded in Storm Events as-reported, so we adjusted all property damages for inflation using monthly consumer price index (CPI) values to Jan 1, 2018 dollars.

Due to changes in reporting frequency and procedures, we did not consider any tornadoes occurring before 1997. These changes include the increasing frequency of EF/F0-EF/F1 (weak) tornadoes reported since 1990 due to scientific advancements and the incorporation of more storm spotters, the 1994 change from reporting tornado path width as an mean of the path to the maximum of the path, and the 1996 change in differentiating between crop and property damage (McCarthy, 2003). The most recent tornado considered here occurred on December 31, 2018. This study period is similar to other recent atmospheric work (Haberlie and Ashley, 2019).

#### *2.1.2. Land cover data*

For each tornado event, we performed a weighted extraction of land cover classes from the MRLC National Land Cover Database (hereafter "NLCD") (Homer et al., 2007, 2015; Fry et al., 2011). This weighted extraction assigns higher weights to the spatial origin of a tornado, and lower weights to regions distant from the origin. The NLCD is not updated annually, so we used the 2001 land cover classes for all tornado events occurring before 2001, the 2006 classes for events occurring between 2006 and 2011, and the 2011 classes for events occurring after 2011.

The weighted extraction was performed using a Gaussian filter centered at the beginning latitude and longitude of the tornado and whose size was determined by the sample standard deviation of tornado path length (sample standard deviation approximately 9054 m; extraction square with approximately 54,330-m side length). From the resulting list of land cover classifications, we calculated the proportion of each classification within the buffer radius weighted by its associated value from the Gaussian filter. We then omitted the "Unclassified" classification and the "Perennial Ice/Snow" classification - which occurred in 38 of 22,123 instances.

#### *2.1.3. Socioeconomic data*

To capture relevant socioeconomic and demographic factors that might explain tornado damages, we acquired 64 variables from the American Community Survey's (ACS) (U.S. Census Bureau; American Community Survey, 2009) five-year estimates (via the tidycensus R package, https://cran.r-project.org/web/packages/tidycensus/index.html). The variables are listed in Table 1 and include demographics, median home value, percent mobile homes, and median home age. The 64 acquired variables resulted in 21 model predictors due to the following aggregations which were performed to yield a smaller set of more informative predictors:

- Number of people with high school education or equivalent, 25 years or older (2 → 1)
  - Males with high school education or equivalent, 25 years or older
  - Females with high school education or equivalent, 25 years or older
- Number of people with associate's degree, 25 years or older (2 → 1)
  - Males with associate's degree, 25 years or older
  - Females with associate's degree, 25 years or older
- Number of people with bachelor's degree, 25 years or older (2 → 1)
  - Males with bachelor's degree, 25 years or older
  - Females with bachelor's degree, 25 years or older
- Number of people with graduate degree, 25 years or older (6 → 1)
  - Males with master's degree, 25 years or older
  - Males with professional school degree, 25 years or older
  - Males with doctorate degree, 25 years or older
  - Females with master's degree, 25 years or older
  - Females with professional school degree, 25 years or older
  - Females with doctorate degree, 25 years or older
- Number of people 65 years or older (12 → 1)
  - Males aged 65-66
  - Males aged 67-69
  - Males aged 70-74
  - Males aged 75-79
  - Males aged 80-84
  - Males aged 85 or over
  - Females aged 65-66
  - Females aged 67-69
  - Females aged 70-74
  - Females aged 75-79
  - Females aged 80-84
  - Females aged 85 or over
- Number of people not working, 16 years or older (20 → 1)
  - Never-married males in the labor force - unemployed, 16 years or

**Table 1**

All variables used in the predictive modeling, their sources, their transformations, and, if applicable, their literature justifications (which either suggest a hypothesized connection between the variable and tornado impacts or strength and/or explicitly uses the variable to explain tornado impacts or strength). "Property Damage" contains an asterisk (*) to indicate that it is the outcome variable, while double asterisks (**) indicate derived variables. **Bolded** variable names indicate variables which can be known given only a location; i.e., without tornado/storm information. "Total Income Estimate for Tornado Area" is *italicized* to indicate that it is a mixture of location and tornado/storm information and, therefore, only used in combined and no-land-cover models, which used all variables and used all variables excluding those from the National Land Cover Database, respectively. A variable name followed by (bs) indicates that it was represented by a cubic B-spline expansion. The transformation column contains a number referencing the transformation equation used on that variable before model fitting, which are

$$1.\ \boldsymbol{x}_{new} = \frac{\boldsymbol{x} - \overline{\boldsymbol{x}}}{s_{\boldsymbol{x}}},$$

$$2.\ \boldsymbol{x}_{new} = \frac{\log(\boldsymbol{x}+1) - \overline{\log(\boldsymbol{x}+1)}}{s_{\log(\boldsymbol{x}+1)}},$$

$$3.\ \boldsymbol{x}_{new} = \frac{\log(1000^*\boldsymbol{x}+1) - \overline{\log(1000^*\boldsymbol{x}+1)}}{s_{\log(1000*\boldsymbol{x}+1)}}.$$

Here $\overline{x}$ represents the sample mean of all $x$ values, $\log(x)$ represents the natural log of all $x$ values, and $s_x$ represents the sample standard deviation of all $x$ values. Notably, the literature justifications for "Beginning Latitude/Longitude" suggest that geographic information is relevant to the occurrence of significant tornadoes and/or tornado impacts, they do not explicitly support the usefulness of *beginning* coordinates.

| Data Source | Variable Name | Transformation | Literature |
|---|---|---|---|
| Storm Events | Property Damage* | 2 | |
| | Tornado Duration | 2 | |
| | Beginning Latitude | 1 | (Ashley and Strader, 2016; Walker, 2007; Concannon et al., 2000; Brooks et al., 2003; The National Severe Storms Laboratory; Simmons and Sutter, 2010) |
| | Beginning Longitude | 1 | (Ashley and Strader, 2016; Walker, 2007; Concannon et al., 2000; Brooks et al., 2003; The National Severe Storms Laboratory; Simmons and Sutter, 2010) |
| | Tornado Length | 2 | Simmons and Sutter (2014) |
| | Tornado Width | 2 | |
| | Tornado Area | 2 | |
| | Multi-Vortex Indicator | | (Prediction Center) |
| | Beginning Time of Tornado Event (bs) | 2 | (Kellner and Dev, 2014; Walker, 2007; Hagemeyer and Schmocker, 1991; Simmons and Sutter, 2010, 2014) |
| | Year of Tornado Event | 1 | (Kellner and Dev, 2014; Brooks et al., 2003; Simmons and Sutter, 2008) |
| | Day of the Year of Tornado Event (bs) | 3 | (Kellner and Dev, 2014; Walker, 2007; Concannon et al., 2000; Brooks et al., 2003; The National Severe Storms Laboratory; Hagemeyer and Schmocker, 1991; Simmons and Sutter, 2014) |
| NLCD | **Open Water Proportion** | 3 | |
| | **Developed Open Space Proportion** | 3 | (Kellner and Dev, 2014; Ashley and Strader, 2016; Ashley et al., 2014; Strader et al., 2016) |
| | **Developed Low Intensity Proportion** | 3 | (Kellner and Dev, 2014; Ashley and Strader, 2016; Ashley et al., 2014; Strader et al., 2016) |
| | **Developed Medium Intensity Proportion** | 3 | (Kellner and Dev, 2014; Ashley and Strader, 2016; Ashley et al., 2014; Strader et al., 2016) |
| | **Developed High Intensity Proportion** | 3 | (Kellner and Dev, 2014; Ashley and Strader, 2016; Ashley et al., 2014; Strader et al., 2016) |
| | **Barren Land Proportion** | 3 | |
| | **Deciduous Forest Proportion** | 3 | Kellner and Dev, 2014 |
| | **Evergreen Forest Proportion** | 3 | Kellner and Dev, 2014 |
| | **Mixed Forest Proportion** | 3 | Kellner and Dev, 2014 |
| | **Shrub/Scrub Proportion** | 3 | |
| | **Pasture/Hay Proportion** | 3 | |
| | **Cultivated Crops Proportion** | 3 | |
| | **Woody Wetland Proportion** | 3 | Kellner and Dev, 2014 |
| | **Emergent Herbaceous Wetland Proportion** | 3 | |
| | **Total Developed Intensity**** | 3 | (Kellner and Dev, 2014; Ashley and Strader, 2016; Strader et al., 2016) |
| | **Total Wooded Proportion**** | 3 | Kellner and Dev, 2014 |
| | **Total Wooded-Developed Interaction**** | 3 | |

(*continued on next page*)

**Table 1** (*continued*)

| Data Source | Variable Name | Transformation | Literature |
|---|---|---|---|
| ACS | **Median Household Income** | 3 | (Kunkel et al., 1999; Changnon et al., 2000; Stimers and Paul, 2016; Simmons and Sutter, 2014) |
| | **Percent Homes that are Mobile**** | 1 | (American Meteorological Society; Walker, 2007; Stimers and Paul, 2016; Simmons and Sutter, 2007, 2008, 2010, 2014) |
| | **Population** | 2 | (Kunkel et al., 1999; Changnon et al., 2000; Walker, 2007; Simmons and Sutter, 2014) |
| | **Median Year Structure Built** | 2 | (Simmons and Sutter, 2008, 2014) |
| | **Number of Homes** | 2 | Stimers and Paul (2016) |
| | **Percent of Pop. that are White**** | 1 | (Strader and Ashley, 2018; Stimers and Paul, 2016; Simmons and Sutter, 2014) |
| | **Percent of Pop. that are Male**** | 1 | (Strader and Ashley, 2018; Stimers and Paul, 2016; Simmons and Sutter, 2014) |
| | **Percent of Pop. that are under 18 years old**** | 1 | (Strader and Ashley, 2018; Simmons and Sutter, 2014) |
| | **Percent of Adults that have High School Education**** | 1 | (Strader and Ashley, 2018; Stimers and Paul, 2016; Simmons and Sutter, 2014) |
| | **Percent of Adults that have Associates**** | 1 | Simmons and Sutter (2014) |
| | **Percent of Adults that have Bachelors**** | 1 | Simmons and Sutter (2014) |
| | **Percent of Adults that have Graduate**** | 1 | Simmons and Sutter (2014) |
| | **Percent of Pop. that are over 65 years old**** | 1 | (Strader and Ashley, 2018; Simmons and Sutter, 2014) |
| | **Lower Quartile Home Value** | 2 | |
| | **Median Home Value** | 2 | (Hall and Ashley, 2008; Stimers and Paul, 2016; Simmons and Sutter, 2014) |
| | **Upper Quartile Home Value** | 2 | |
| | **Percent of Pop. Experienced Poverty Last 12 Months**** | 2 | (Strader and Ashley, 2018; Stimers and Paul, 2016) |
| | **Gini Index** | 2 | Simmons and Sutter (2014) |
| | **Percent of Adults not Working**** | 1 | (Strader and Ashley, 2018; Stimers and Paul, 2016) |
| | **Percent of Adults that Commute over 30min**** | 1 | Simmons and Sutter (2014) |
| | **Percent of Adults that Depart between 00:00 and 04:59**** | 1 | |
| Storm Events, ACS | *Total Income Estimate for Tornado Area* | 1 | (Kunkel et al., 1999; Changnon et al., 2000) |

older
- – Never-married males not in the labor force, 16 years or older
- – Never-married females in the labor force - unemployed, 16 years or older
- – Never-married females not in the labor force, 16 years or older
- – Married males in the labor force - unemployed, 16 years or older
- – Married males not in the labor force, 16 years or older
- – Married females in the labor force - unemployed, 16 years or older
- – Married females not in the labor force, 16 years or older
- – Separated males in the labor force - unemployed, 16 years or older
- – Separated males not in the labor force, 16 years or older
- – Separated females in the labor force - unemployed, 16 years or older
- – Separated females not in the labor force, 16 years or older
- – Widowed males in the labor force - unemployed, 16 years or older
- – Widowed males not in the labor force, 16 years or older
- – Widowed females in the labor force - unemployed, 16 years or older
- – Widowed females not in the labor force, 16 years or older
- – Divorced males in the labor force - unemployed, 16 years or older
- – Divorced males not in the labor force, 16 years or older
- – Divorced females in the labor force - unemployed, 16 years or older
- – Divorced females not in the labor force, 16 years or older

- People who commute over 30 min to work, workers 16 years or over who did not work at home (6 → 1)
- People who commute 30–34 min to work, workers 16 years or over who did not work at home
- People who commute 35–39 min to work, workers 16 years or over who did not work at home
- People who commute 40–44 min to work, workers 16 years or over who did not work at home
- People who commute 45–59 min to work, workers 16 years or over who did not work at home
- People who commute 60–89 min to work, workers 16 years or over who did not work at home
- People who commute 90 or more minutes to work, workers 16 years or over who did not work at home

These socioeconomic and demographic factors were incorporated with tornadoes by using the same Gaussian template used in the NLCD methodology, such that multiple counties' information could be incorporated.

Data with associated county shapes are not available before 2010 nor after 2017, so we assumed values before 2010 to be equal to the 2010 values, while values after 2017 were equal to the 2017 values. Events occurring for years with ACS data were matched with their same year.

During the automated incorporation of ACS variables, some missing predictor values occurred, which resulted in the omission of 75 out of 22,123 tornado events from the analysis. A map of these omitted tornadoes' beginning coordinates is displayed in Fig. 2 and a comparison of omitted versus retained reported property damages are displayed in Fig. 3. Due to their distribution and small amount, we do not expect the omission of these events to systematically bias the results. Manual inspection confirmed that at least 67 (89%) of these were due to missing estimates from ACS.

### 2.2. *Derived explanatory variables*

After all data sources were integrated into the data set, we produced several derived explanatory variables that we expected to be predictive of tornado damages (in addition to the composite variables listed under section 2.1.3). "Tornado Area" is the product of "Tornado Length" and "Tornado Width". Multi-vortex tornadoes tend to cause more severe damage (Prediction Center), so we performed a text-search on the Storm Event's "event narratives" to determine whether a tornado was a multi-vortex tornado; this is represented in the binary "Multi-Vortex

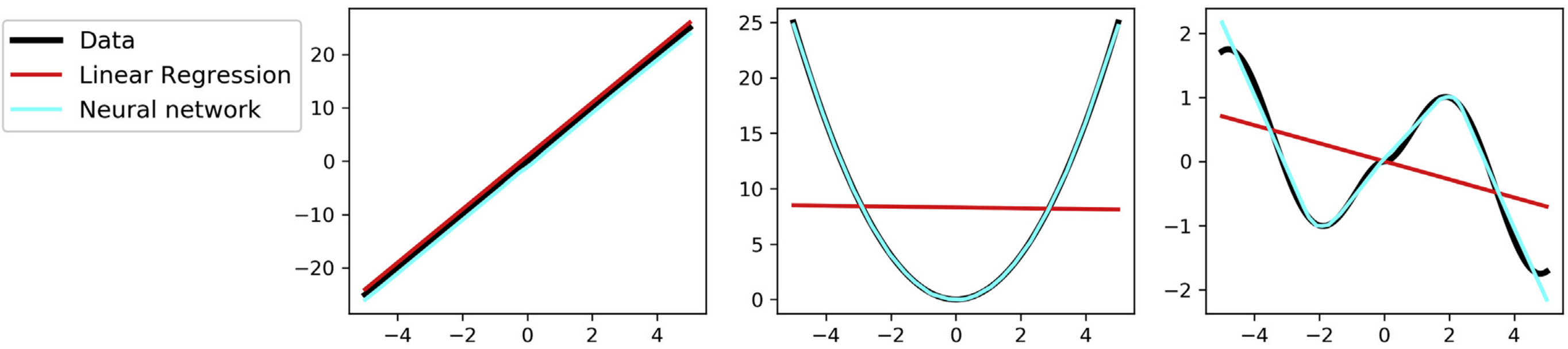

**Fig. 1.** A demonstration of the pattern learning abilities of artificial neural networks relative to simple linear regression. Black lines are true data, red lines are linear regression estimates, and cyan lines are artificial neural network estimates. This is displayed for a linear ($y = 5x$), quadratic ($y = x^2$), and complex case ($y = \sin(x)\log(|x| + 1)$). The artificial neural network had one hidden layer consisting of 32 hidden units, ReLU activation functions, and trained via the Adam optimization method. The linear case lines are slightly shifted to prevent complete overlap. (For interpretation of the references to color in this figure legend, the reader is referred to the Web version of this article.)

Indicator" variable. "Beginning Time" and "Day of the Year" are both represented as basis expansions using cubic splines with 8 and 12 evenly spaced knots over minutes since midnight and day of the year, respectively, to account for non-linear time effects.

"Total Developed Intensity" is the sum of each "Developed" land cover proportion multiplied by the median value of that NLCD classification's impervious surface cover. "Total Wooded Proportion" is the sum of the "Woody" and "Forest" classification proportions, and "Total Wooded-Developed Interaction" is the product of "Total Developed Intensity" and "Total Wooded Proportion".

ACS variables for number of people whose race indicates "white", number of males, number of people under 18 years old, number of people over 65 years old, and number of people who experienced poverty in the last 12 months were all divided by the total population to derive percentages. Additionally, ACS variables for people with various education levels (high school, associate's, bachelor's, and graduate), number of people unemployed or not in the labor force, number of people who commute over 30 min to work, and number of people who leave home for work between the hours of 12:00 A.M. and 4:59 A.M. were all divided by the total number of people over the age of 18 to derive approximate percentages (approximate because some variables state relevance to people over 16, 25, etc…). Lastly, number of mobile homes was divided by total number of housing units to derive a percentage of homes that were mobile, and "Tornado Area" and "Median Household Income" are multiplied to derive "Total Income Estimate for Tornado Area". All variables along with their data source(s), transformation method, and literature justification(s), if applicable, are named in Table 1.

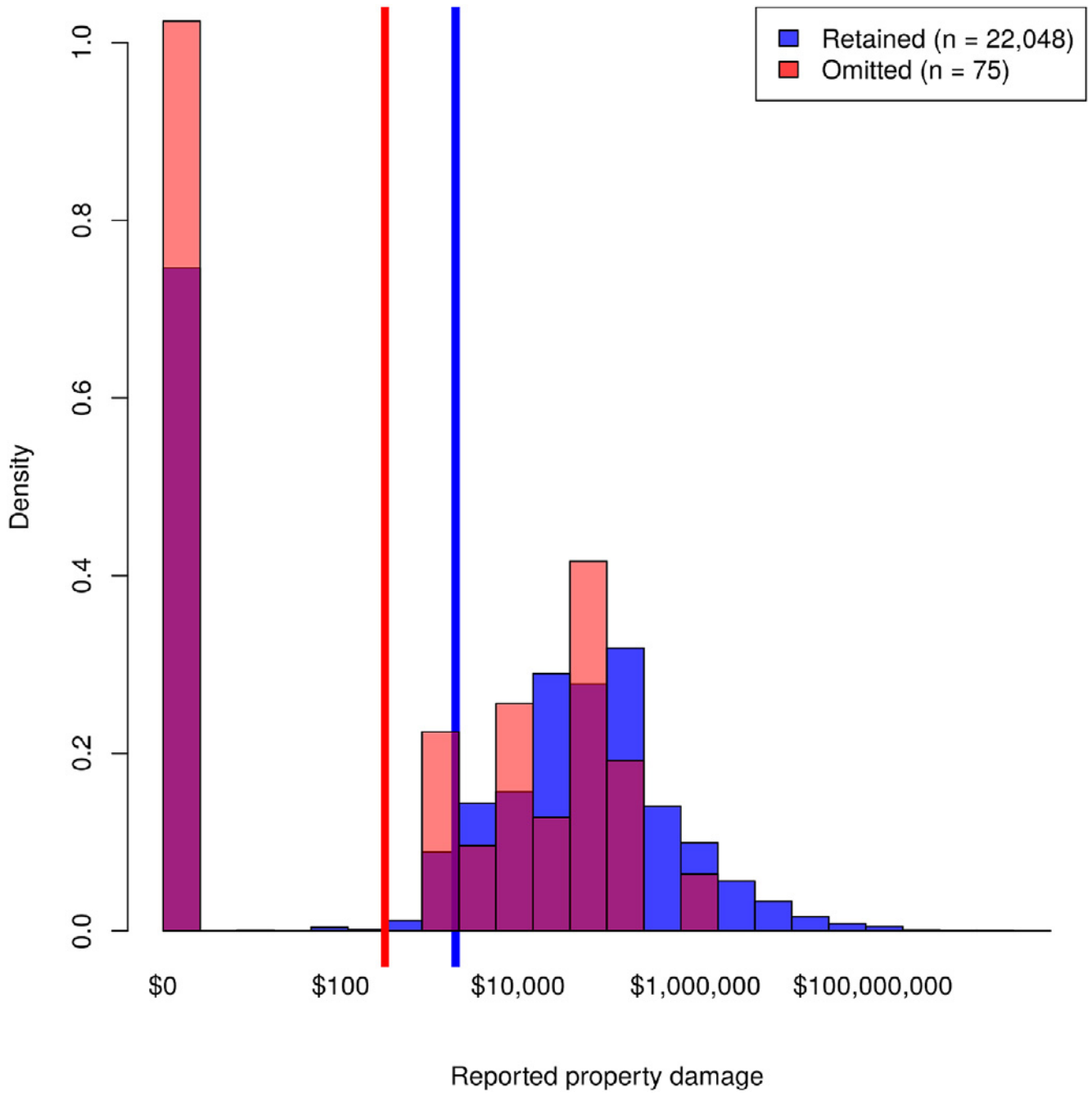

**Fig. 3.** The distribution of retained (blue, n = 22,048) and omitted (red, n = 75) tornadoes are highly similar, with no particularly extreme tornadoes being omitted. The red vertical line locates the sample mean for the omitted tornadoes while the blue vertical line locates the sample mean for the retained tornadoes. (For interpretation of the references to color in this figure legend, the reader is referred to the Web version of this article.)

### *2.3. Data processing and handling*

To avoid ill-conditioning and to stabilize parameter learning, all variables were processed to have mean zero and unit standard deviation. To promote trend learning, variables with high variance were preprocessed with log transformations to further centralize distributions. All transformation equations are provided in Table 1.

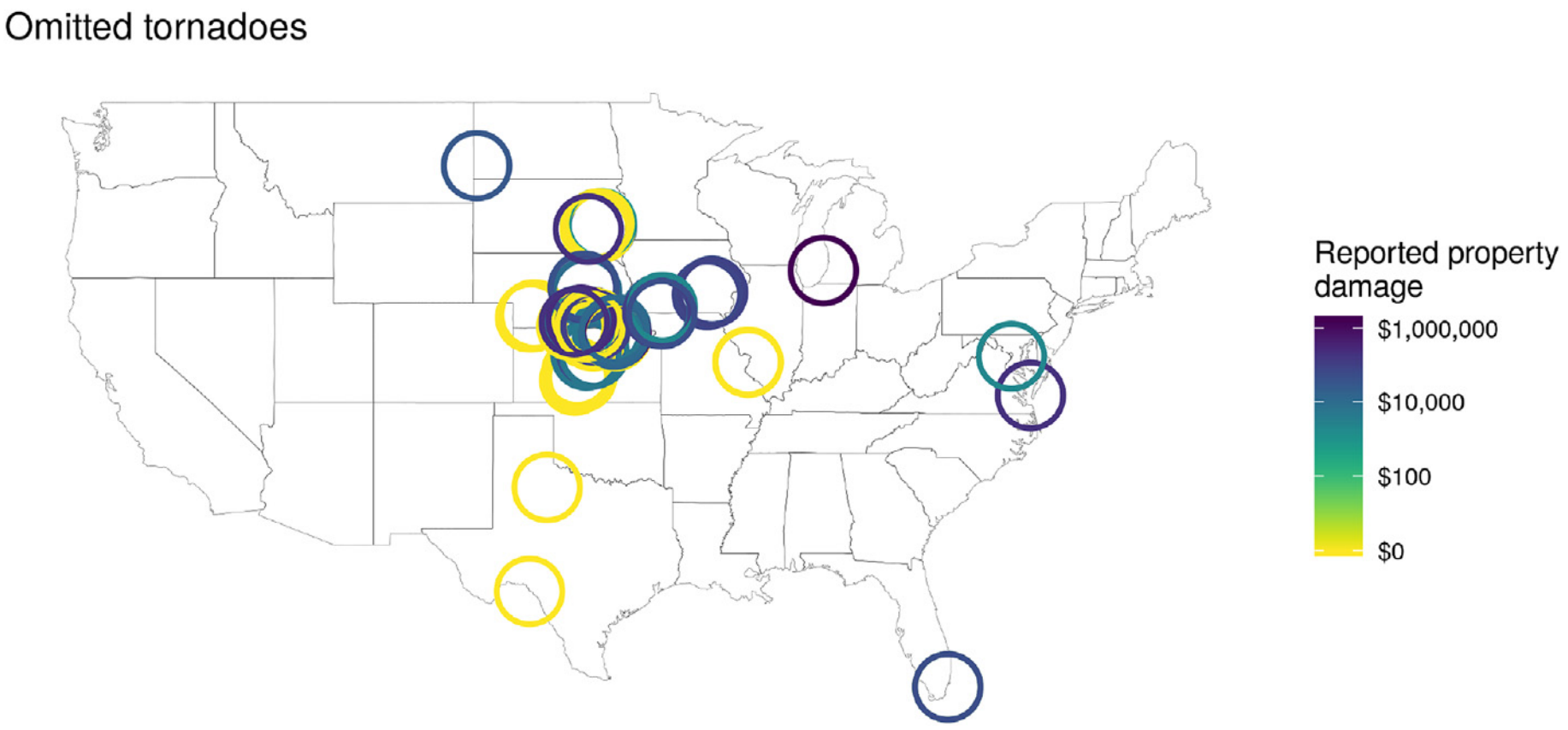

**Fig. 2.** A map of tornadoes which were omitted (n = 75) from the data set due to the emergence of NA values when incorporating American Community Survey (ACS) data via Gaussian-weighted extraction. Color represents damages as-reported (i.e. not adjusted for inflation). (For interpretation of the references to color in this figure legend, the reader is referred to the Web version of this article.)

After processing, the data were randomly partitioned into three sets: a training set (consisting of 60% of events), cross-validation set (20%), and test set (20%). The training set was used to optimize the model coefficients/parameters, while the cross-validation set was used to determine the best model among those of differing hyperparameters (such as neural network architecture and regularization strength). After the best hyperparameter-defined model was found, that model was retrained on both the training set and the cross-validation set (combined 80% of the data), and the test set was used to estimate the best model's predictive performance. After that, the model was retrained on the entire data set to create a final model to generate future predictions. At each stage, predictive performance was measured as mean squared error (MSE) between predicted and true property damage values.

## 2.4. Models

### 2.4.1. Brief introduction to artificial neural networks

This subsection serves to provide a very brief explanation and demonstration of the type of neural networks used in this study. For a more thorough discussion of neural networks as they related to Earth system science, see (Reichstein et al., 2019) - while (Hastie et al., 2009) provides a more mathematical approach.

Neural networks learn (via optimization of the error/loss function) intermediate variables (hidden units/neurons) which are linear combinations of the previous features (either input variables or other hidden units) but then transformed by non-linear functions (activation functions), such as the logistic, hyperbolic tangent, or rectified linear unit functions. After some number of hidden layers (each consisting of hidden units), the linear combination of previous variables may be mapped to a target variable via an identity function (continuous case) or a sigmoid function (binary case), among others.

Considering the case of three input variables ($x_1$, $x_2$, $x_3$) and one output variable ($y$), a neural network with one hidden layer with two hidden units could be written as:

$$\hat{y} = I(\beta_{1,1} f(\alpha_{1,1}x_1 + \alpha_{1,2}x_2 + \alpha_{1,3}x_3) + \beta_{1,2} f(\alpha_{2,1}x_1 + \alpha_{2,2}x_2 + \alpha_{2,3}x_3)),$$

where $\alpha_{i,j}$ is the coefficient/weight mapping the $j^{th}$ input variable to the $i^{th}$ hidden unit, $\beta_{1,j}$ is the coefficient mapping the $j^{th}$ hidden unit to the target estimate $\hat{y}$, $f()$ represents the activation function, and $I()$ represents the identity function. The addition of a second hidden layer with two hidden units changes the equation to:

$$\hat{y} = I(\lambda_{1,1} f(\beta_{1,1} f(\alpha_{1,1}x_1 + \alpha_{1,2}x_2 + \alpha_{1,3}x_3) + \beta_{1,2} f(\alpha_{2,1}x_1 + \alpha_{2,2}x_2 + \alpha_{2,3}x_3)) + \lambda_{1,2} f(\beta_{2,1} f(\alpha_{1,1}x_1 + \alpha_{1,2}x_2 + \alpha_{1,3}x_3) + \beta_{2,2} f(\alpha_{2,1}x_1 + \alpha_{2,2}x_2 + \alpha_{2,3}x_3))),$$

where $\beta_{i,j}$ maps the $j^{th}$ first-layer hidden unit to the $i^{th}$ second-layer hidden unit and $\lambda_{1,j}$ maps the $j^{th}$ second-layer hidden unit to the target estimate.

To demonstrate the robust pattern-fitting ability of neural networks relative to simple linear regression, we sampled 10,000 values from a uniform distribution bound by $-5$ and 5, then simulated observations from three deterministic models, and trained a neural network (1 hidden layer, 32 hidden units, ReLU activation functions, Adam optimization) and linear regression to estimate the observed data set. The deterministic models are as follows:

1. $y = 5x$

2. $y = x^2$

3. $y = sin(x)log(|x| + 1)$

The results of this are displayed in Fig. 1.

All neural networks were made in the PyTorch deep learning framework (http://pytorch.org/). These models were produced using the mini-batch gradient descent optimization algorithm with a batch size of 50, along with the AdaGrad parameter-updating method (Duchi et al., 2011), and rectified linear unit (ReLU) activation functions (except for the output layers, which had an identity function) unless stated otherwise.

### 2.4.2. Continuous property damage models

We initially considered five variable-set models: (1) "beforehand" models excluded all variables regarding the tornado event, these variables are shown in bold font in Table 1; (2) "storm characteristic" models excluded all variables regarding the location of the tornado (such as home values and land cover proportions), these variables are shown in normal font in Table 1; (3) "combined models" used all the variables contained in both (1) and (2) with the addition of the "Total Income Estimate for Tornado Area" variable (which includes both location and tornado information); (4) "no LC models" used all variables except those utilizing NLCD variables (representing the physical environment of the affected area); and, (5) "no ACS models" used all variables except those utilizing ACS variables (representing the socioeconomic and demographic status of the affected area).

For each model type, we developed several artificial neural networks. Artificial neural networks were made with a two-thirds descending number of neurons per hidden layer until a layer contained only 4 neurons (if a descent led to less than 4 neurons, it was rounded up to 4), and this layer was then connected to a one-neuron output layer. Non-integer values produced by the descent were rounded to the nearest integer, and that new integer then was used in determining the subsequent hidden layer. For each possible hidden layer under this two-thirds descending rule, we also created a model which maps that hidden layer directly to an output layer. For example, if a model had nine input variables, we would make (1) a one-hidden-layer neural network with six neurons and (2) a two-hidden-layer neural network with six neurons in the first hidden layer and four hidden neurons in the second/final hidden layer.

This same approach was repeated, omitting tornado events which caused no property damage, to provide an evaluation of the predictive performance conditioned on the premise that a tornado did cause damage (hereafter "conditional models").

We then explored additional neural network architectures on the combined models. Rather than exclusively descending model architectures, as previously described, we developed models that were limited to 2 hidden layers but with a variable number of neurons ("wide models") and models that were limited to 34 neurons (the result of dividing the number of input variables by two) but with variable number of hidden layers ("deep models").

Wide models with 20 or more neurons per hidden layer displayed overfitting (noticeably better performance on the training set relative to the cross-validation set). Thus, we implemented dropout regularization (Srivastava et al., 2014) on a wide model with 100 neurons to allow for penalized complexity. Dropout regularization uses a predefined probability to independently and randomly set each hidden unit to zero during parameter optimization. The dropout probability is an additional hyperparameter that was varied from 0.1 to 0.9 by 0.1 increments with the goal of choosing an optimal value. Additionally, we evaluated the performance of models regularized by a combination of L2 and dropout regularization with the L2 regularization strength varied by a factor of 10 from 0.0001 to 100.

These same three combined model architectures (descending, deep, and wide) were then tested on models using the exponential linear unit activation function (ELU) (Clevert et al., 2016) in place of ReLU. Wide ELU models were less prone to overfitting than wide ReLU models, so only dropout regularization was implemented.

### 2.4.3. Binary damage occurrence models

In addition to the conditional models, we developed wide neural networks to predict whether a tornado will cause damage (hereafter "damage classifiers"). These wide neural networks were created using the same scheme described for the previous models except that the one-neuron output layer used a logistic activation function, providing a probability of a tornado causing damage. These models were optimized using binary cross entropy and evaluated by the area under the receiver operating characteristic curve (AUROC). Again, wide models with many neurons displayed overfitting, and we implemented dropout regularization.

### 2.4.4. Zero-inflated log-normal regression

To provide a more traditional comparison to the neural networks, we initially set out to fit zero-inflated Poisson and zero-inflated negative binomial regressions; however, due to the log transformation and adjustments for inflation, the dollars are not traditional count data. So, we instead used a zero-inflated log-normal distribution, which has been used in several other zero-inflated semi-continuous cases (Calama et al., 2011; Mayer et al., 2005; Belasco and Ghosh, 2008; Rossen et al., 2013; Ning et al., 2011).

In brief, the zero-inflated log-normal regression attempts to model whether the variable is equal to 0 with a logistic regression, and given that the variable exceeds 0, it models the log transformation of the variable with a normal regression. These models appropriately apply when the response variable has a large number of zeros, cannot be negative, and has a small amount of very large values (Calama et al., 2011).

With the binary damage occurrence being modeled by a logistic regression and the (semi-)continuous property damage being modeled by a log-normal, this approach is highly comparable to our neural networks with the exclusion of hidden layers and the regularization and optimization methods which that increased complexity necessitates. Due to the similarities in use case and actual mathematics, we refer to our modeling approach as "zero-inflated neural networks".

### 2.5. Future predictions and dashboard

To generate new predictions, we created a rectangular grid of points uniformly spaced by 0.75° from −125 to −66° longitude and 23–50° latitude. A census-provided U.S. boundary was used to remove any prediction points not within the contiguous United States. All physical and social variables were determined for these points using the same methodology used in the original data set fusions, and mean values were assigned for storm characteristic variables.

This predictor gathering process was also done for 261 US cities that had a 2014 population exceeding 100,000.

Using the best conditional model architecture and damage classifier (both trained on 100% of the data set), we computed predictions for all cities and grid points for the 15th of every 2019 month. These predictions are highlighted in-print, with increased emphasis on areas prone to tornado research, and online via an interactive dashboard.

All code files and notebooks to reproduce this work (data handling, analysis, and visualization) are publicly available at https://github.com/jdiaz4302/tornadoesr/, and, all the necessary data for replication and innovation is publicly available at (Diaz and Joseph, 2018).

## 3. Results

### 3.1. Continuous property damage models

Of our original 5 variable sets (beforehand, storm characteristic, combined, No LC, and No ACS), models which excluded non-damaging tornadoes (predicting $Y|Y > 0$, "conditional") consistently had lower MSE on out-of-sample data than those which included them (predicting $Y|Y \geq 0$), as displayed in Fig. 4.

Overall, combined models, which used all available variables without regularization, provided the lowest cross-validation set MSE (MSE = 0.0955, $R^2$ = 0.440). Conditional beforehand models, which included land cover, socioeconomic, and demographic variables, achieved a maximum $R^2$ of 0.031. No ACS models, which omitted socioeconomic and demographic variables, had better cross-validation set performance (MSE = 0.0973, $R^2$ = 0.430) than No LC models (MSE = 0.102, $R^2$ = 0.405), which omitted land cover variables. The performance for each architecture type for each variable set, excluding and including non-damaging tornadoes is shown in Fig. 5.

When descending, wide, and deep neural network architectures were explored for combined conditional models, wide models displayed the best performance (MSE = 0.0935, $R^2$ = 0.452). When highly parameterized wide models (100 hidden units in each of 2 hidden layers) were developed with dropout regularization, performance peaked (MSE = 0.0903, $R^2$ = 0.471) at 20% dropout probability. When these same models were developed with exponential linear unit (ELU) activation functions (rather than rectified linear unit), they were able to achieve comparable but ultimately lower performance (MSE = 0.0914, $R^2$ = 0.464). The performance for each architecture variant for these combined conditional models are displayed in Fig. 6.

After the best-performing wide model was trained on both the training and cross-validation set, it was then evaluated on the test set for a final out-of-sample analysis. This yielded a MSE of 0.0918 and a $R^2$ of 0.432. A plot of observed versus predicted test set property damages for this model is displayed in Fig. 7.

### 3.2. Binary damage occurrence models

Of the wide neural networks considered for the binary damage occurrence models, a highly parameterized (100 hidden units in each of 2 hidden layers) and lightly regularized (10% dropout) model performed best (accuracy = 0.839, AUROC = 0.894). Once retrained on both the training and cross-validation set, a final out-of-sample analysis revealed an accuracy of 0.821 and AUROC of 0.872. A plot of outcome versus predicted test set probabilities for this model is displayed in Fig. 8.

### 3.3. Zero-inflated log-normal regression

On the cross-validation tornadoes which did cause damage, the zero-inflated log-normal regression achieved a MSE of 0.100, failing to match the performance of the neural networks. Likewise, this model achieved a cross-validation accuracy of 0.813 and AUROC of 0.853.

### 3.4. Maps

Once the two best neural networks (conditional and binary) were trained on both the training and cross-validation set, the absolute and squared values of test set residuals for the conditional and binary model, respectively, were plotted at the tornadoes' beginning coordinates. For easier visual interpretation, we also used the mean of their bilinear interpolation to display a spatial grid of residuals (Fig. 9).

After the residual maps were created and test set performance was determined, the two best neural networks were trained on the entire data set (training, cross-validation, and test set). From these models, maps were created from predictions of conditional property damage and probability of damage occurring across a US-wide grid with additional prediction points for large U.S. cities. For print, we have provided prediction maps for the $15^{th}$ of the month with the highest predicted conditional damages for the displayed area - contiguous U.S., Kansas, Alabama, Illinois, Oklahoma, and Florida (Fig. 10).

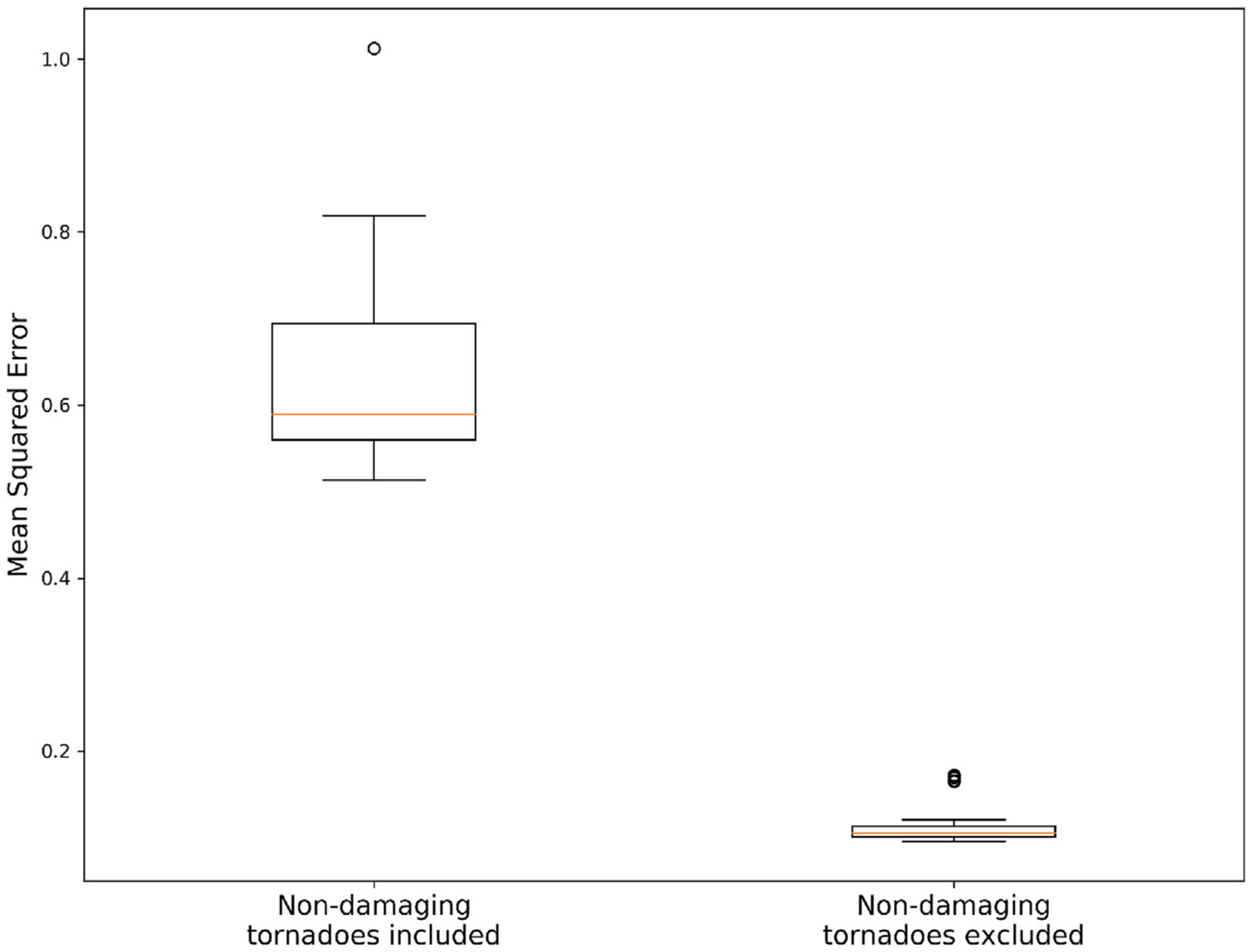

**Fig. 4.** A standard box plot showing the mean squared error (MSE) of models that were trained on the entire data set ("Non-damaging tornadoes included") and a subset of the data set which excluded non-damaging tornadoes ("Non-damaging tornadoes excluded").

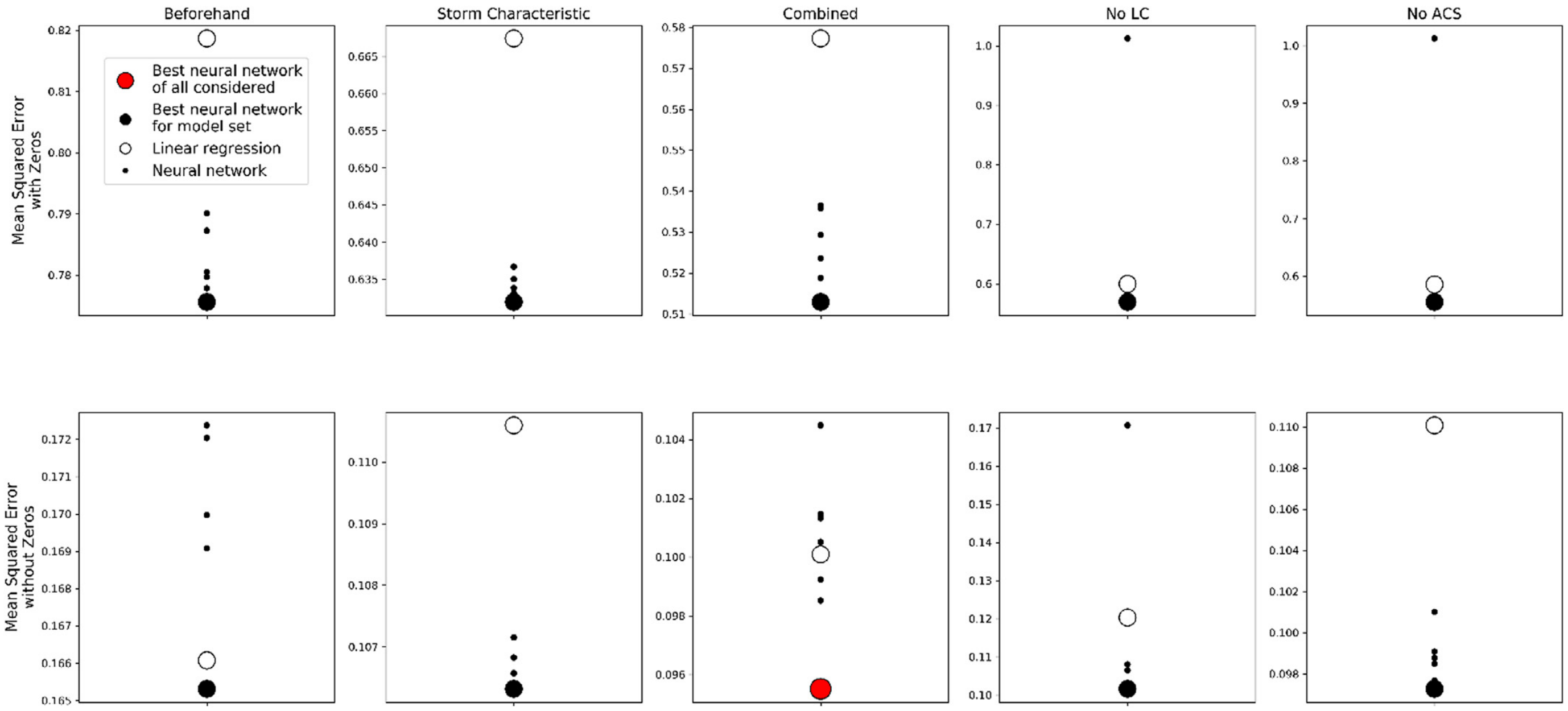

**Fig. 5.** A series of scatter plots, where each plot displays the mean squared error (MSE) on the y-axis associated with column-titled models. Row titles indicate whether the models were trained on the entire data set (with zeros) or on a subset of the data set which excluded non-damaging tornadoes (without zeros). Small black points indicate a neural network, large hollow points indicate the multiple linear regression trained on that variable set, large black points indicate the best neural network of that variable set, and the large red point indicates the best model (lowest MSE) of all displayed. Note the varying y-axis scales. (For interpretation of the references to color in this figure legend, the reader is referred to the Web version of this article.)

## 4. Discussion

### *4.1. Model insights*

We ultimately determined that it was most appropriate to model tornado-induced property damage with two neural network components. This approach highly mimics traditional zero-inflated models, particularly the zero-inflated log-normal regression which models non-zero and non-negative continuous values (i.e. inflation adjusted dollars) via log-normal regression and excess zero values via logistic regression. The primary difference is the presence of hidden layers in the neural networks that facilitate the representation of possibly complex functions relating inputs and outputs, and, as such, we call these models “zero-inflated neural networks”.

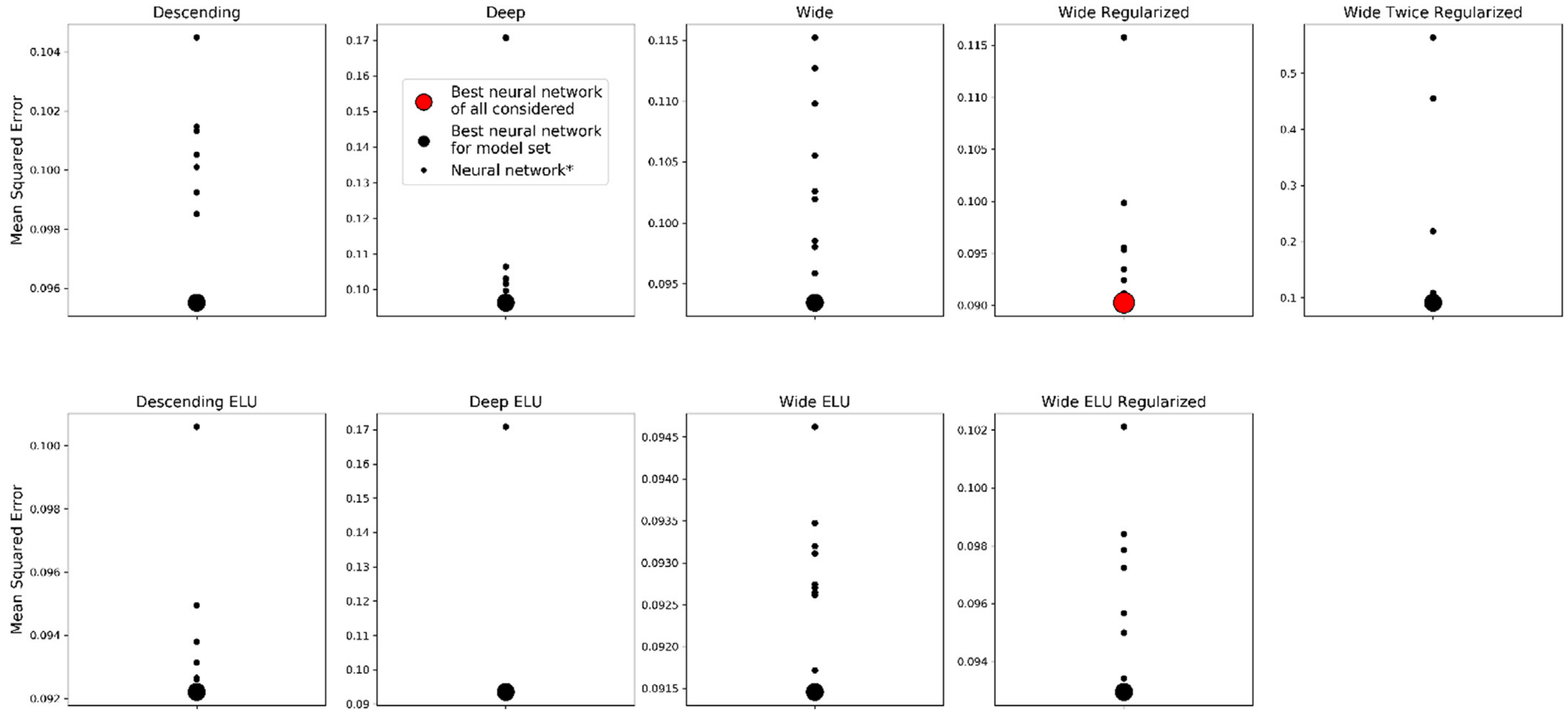

**Fig. 6.** A series of scatter plots, where each plot displays the mean squared error (MSE) on the y-axis associated with column-titled models trained on the non-zero damage data. Small black points indicate a neural network (with the exception of one multivariable linear regression in both of the descending models), large black points indicate the best neural network of that variable set, and the large red point indicates the best model (lowest MSE) of all displayed. (For interpretation of the references to color in this figure legend, the reader is referred to the Web version of this article.)

Of the variable sets that we considered, those which included all variables (storm, land cover, socioeconomic, and demographic) performed best on out-of-sample data. When examined alone, storm variables accounted for most of the performance ability, while the set consisting of only land cover, socioeconomic, and demographic variables performed relatively poorly. Models which omitted land cover variables saw a greater performance loss than those which omitted social variables (socioeconomic and demographic). This would suggest that the order of variable importance for predicting tornado property losses in decreasing order is storm, land cover, then socioeconomic and demographic variables.

Overall, the results of the models that used all input variables were relatively insensitive to hyperparameter values such as the hidden layer and hidden unit scheme, regularization strength, and activation functions. When considering the best models from each hyperparameter exploration, the largest difference in mean squared error was approximately 0.006 and the largest difference in $R^2$ was approximately 0.035. That being said, network architectures with few hidden layers (2) but many hidden units (100) regularized with a small dropout probability ($0 \leq p \leq 20$) performed consistently better than other hidden layer and hidden unit schemes.

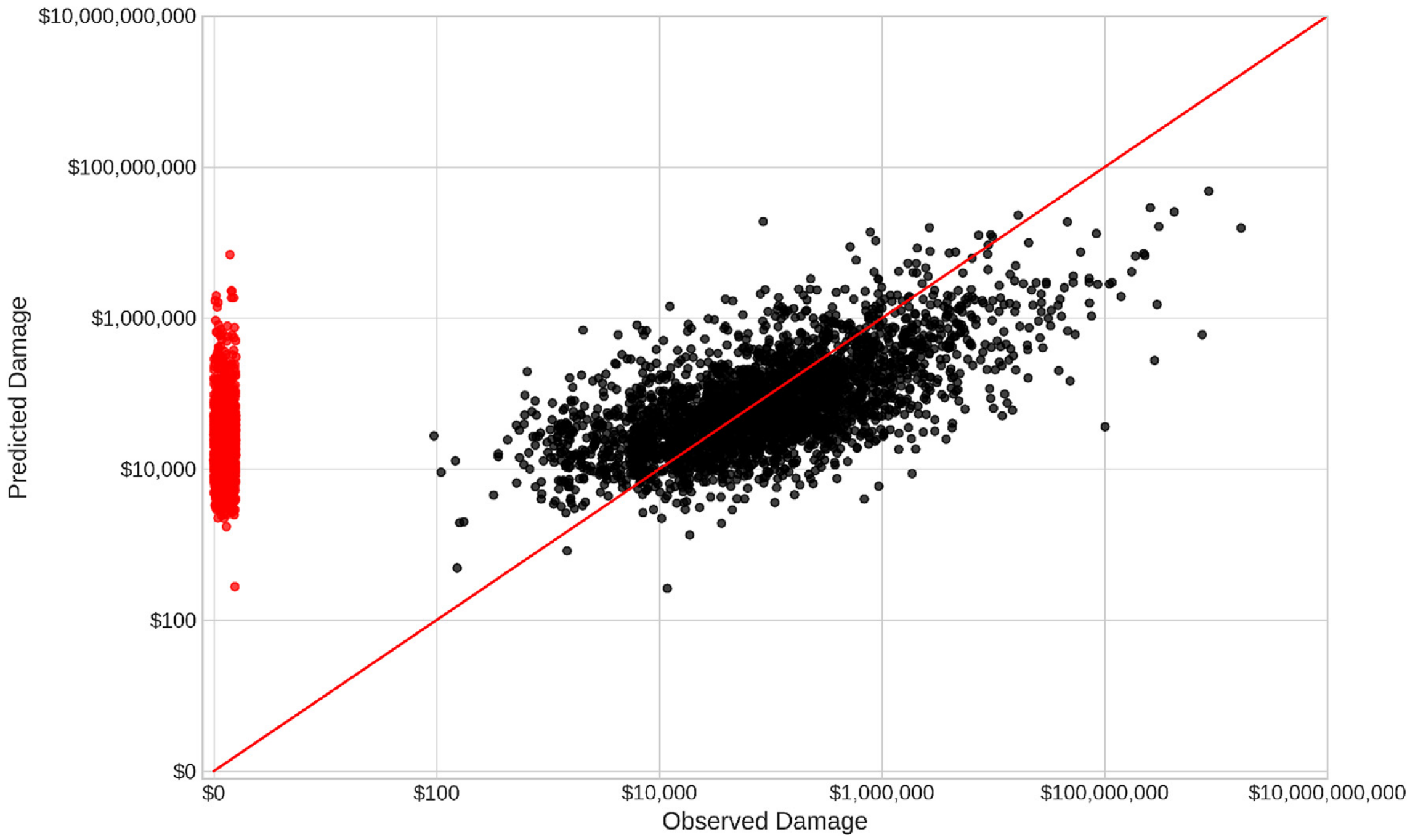

**Fig. 7.** A scatter plot of predicted versus observed property damages for the test set from the best model (lowest mean squared error) for predicting conditional property damage ($Y|Y > 0$). This model was an artificial neural network with 2 hidden layers, 100 hidden units in each hidden layer, rectified linear unit activation functions, 20% dropout regularization, and an identity output function. The diagonal red line indicates a 1:1 relationship (slope = 1, intercept = 0), and the red points indicate tornadoes with an observed $0 property damage. (For interpretation of the references to color in this figure legend, the reader is referred to the Web version of this article.)

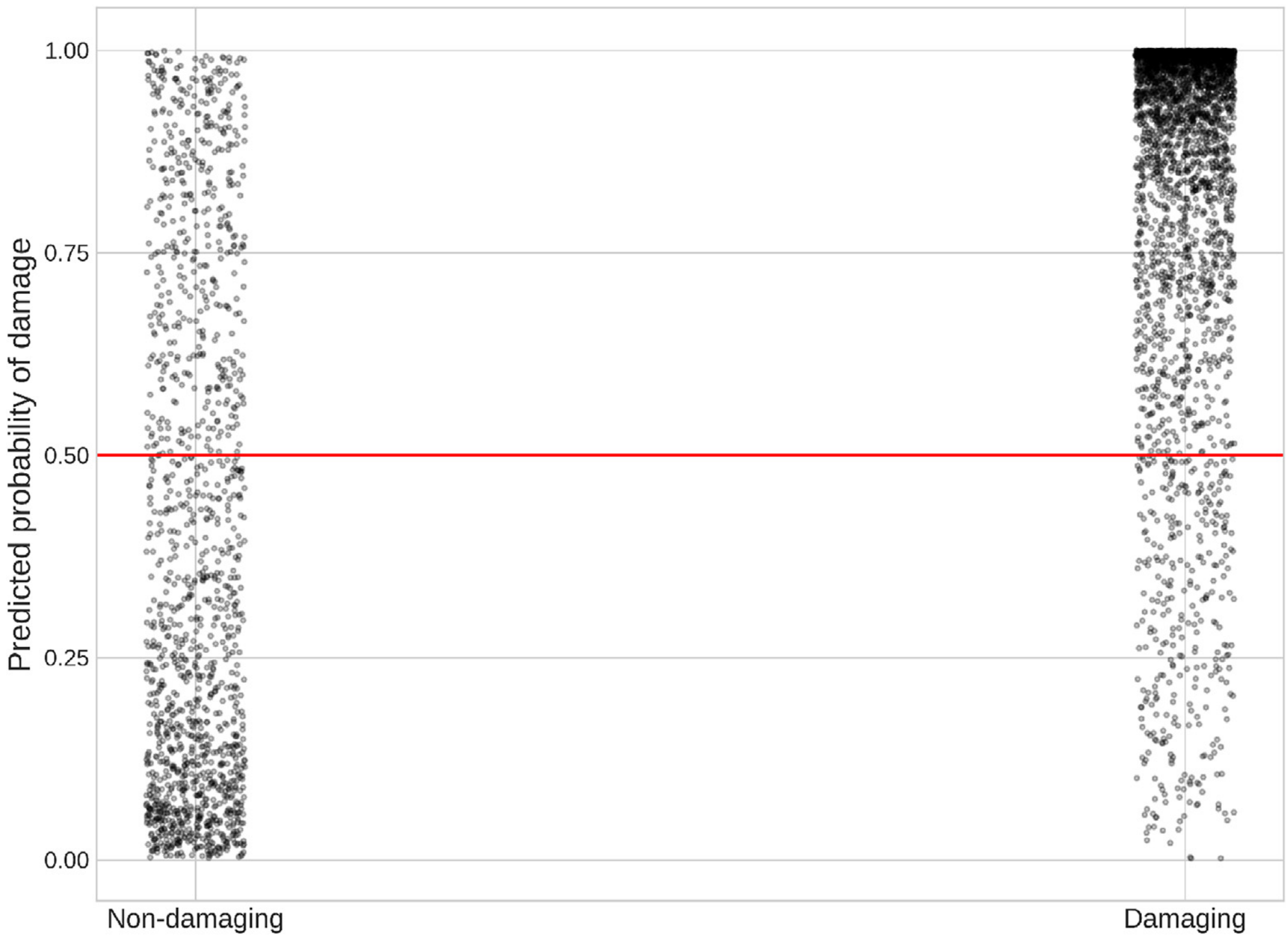

**Fig. 8.** A scatter plot of predicted probability of damage occurrence versus a binary indicator of whether damage occurred for the test set from the best model (lowest area under the receiver operating characteristic curve) for predicting binary property damage occurrence. This model was an artificial neural network with 2 hidden layers, 100 hidden units in each hidden layer, rectified linear unit activation functions, 10% dropout regularization, and a logistic output function. The horizontal red line indicates a probability of 0.50 - the threshold used for converting the probabilities into occurrence predictions in this paper. (For interpretation of the references to color in this figure legend, the reader is referred to the Web version of this article.)

### 4.2. Comparison to other studies

It is common practice for studies to place past tornado paths and intensities onto new locations and then evaluate the area of developed land (Strader et al., 2016; Ashley et al., 2014), number of people (Hall and Ashley, 2008; Rosencrants and Ashley, 2015; Ashley et al., 2014), number of housing units (Hall and Ashley, 2008; Strader et al., 2016; Rosencrants and Ashley, 2015; Ashley et al., 2014), and/or amount of home/property value (Elsner et al., 2018; Hall and Ashley, 2008) exposed to the hypothetical disaster. Previous work has randomly permuted historical disasters to simulate new disasters (Elsner et al., 2018) and/or used a combination of ratios and rates to convert potentially exposed property into dollar amounts of damaged property (Grieser and Terenzi, 2016). However, without external validation - e.g., via withheld data - it is difficult to know how accurate these approaches are for prediction tasks. By evaluating the performance of zero-inflated neural networks on withheld property damage data, we have established that a combination of physical and social features in the spatial neighborhood of a storm's spatial coordinates provide information on subsequent property damage. This raises the question then of how much additional predictive power might be achieved by incorporating explicit modeling of spatial paths, intensity distributions, and/or economic loss models (Grieser and Terenzi, 2016).

Detailed spatially-explicit predictions of tornado-induced property damages are sparse, but a separate study found that mobile homes in Alabama had much higher tornado impact potential than those in Kansas (Strader and Ashley, 2018). Although not focusing solely on mobile homes, our models predicted that city and grid points lying within and adjacent to Alabama had a mean 0.715 probability of damage occurring, while Kansas's corresponding value was 0.636. However, conditioned on damage occurring, Kansas's mean ($138,081) and maximum ($1,134,219) predicted property damage slightly exceeded that of Alabama ($119,043 and $908,927, respectively). In a comparison of three states, it was found that Monte Carlo simulated tornadoes affected a maximum of 198.42, 455.53, and 590.97 housing units in Oklahoma, Alabama, and Illinois, respectively (Strader et al., 2016). Similarly, our models predicted a mean probability of damage occurring of 0.578, 0.715, and 0.807 for Oklahoma, Alabama, and Illinois, respectively. Additionally, mean and maximum conditional property damage predictions were lower for Alabama than Illinois, $187,698 and $1,112,483. Although other tornado impact work found Oklahoma to be among the deadliest tornado states (Ernest Agee and Taylor, 2019), the fact that Oklahoma's mean ($159,561) conditional property damage prediction breaks the impact trend found in (Strader et al., 2016) is likely because of Oklahoma's proximity to the Dallas metropolitan area, which was found to have the greatest potential for a tornado disaster among 4 other metropolitan areas (Rosencrants and Ashley, 2015) - supported by our relatively large $1,411,481 prediction in that area. Additionally, our predictive maps display an increased amount of high probabilities of damage for the American south - an area found by climatological studies to be prone to EF2 (strong) and EF4 (violent) tornadoes (Concannon et al., 2000; Schaefer et al., 1986) as well deadly nocturnal tornadoes (Ashley et al., 2008).

### 4.3. Limitations and future directions

Underestimation of property damages at the upper extremes is a notable obstacle, as the log transformation leads to numerical stability during training, but sacrifices information contained in the natural scale of the response. For example, predicting $27 in damage when the actual value is $10 leads to equivalent squared error loss on the log scale as predicting $367,879 in damage when the actual value is $1,000,000, despite the first prediction being wrong by $17 dollars, and the second prediction being wrong by over $600,000. Future work may benefit from combining a deep learning approach with extreme value approaches that model heavy tails without relying on log transformations, e.g., estimating the parameters of the generalized Pareto distribution with a neural network for exceedances over a threshold amount of damage (Carreau and Bengio, 2009).

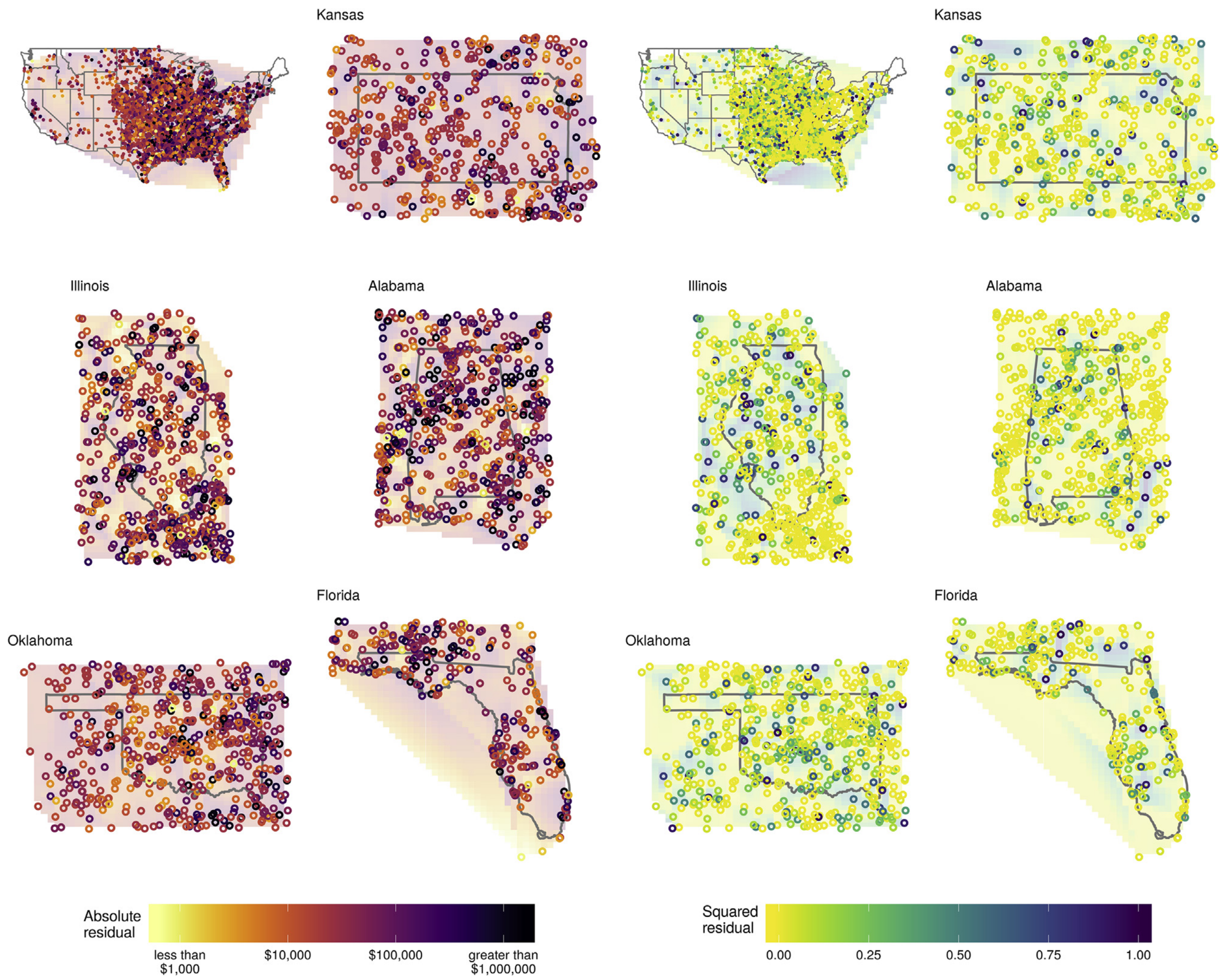

**Fig. 9.** Maps of test set residuals (n = 4410). The two left columns of maps represent absolute residuals for the best conditional property damage model ($Y|Y > 0$), while the two right columns of maps represent the squared residuals for the best property damage occurrence models. Raw point data is plotted over a semi-transparent bilinear interpolation. Darker colors indicate larger residuals. (For interpretation of the references to color in this figure legend, the reader is referred to the Web version of this article.)

The test set residuals are displayed geographically in **Supplement 1** (available at: https://rawgit.com/jdiaz4302/tornadoesr/master/interactive_model_maps.html). This map displays regions prone to over- or under-estimation of property damages, perhaps suggesting the absence of spatially-correlated explanatory variables. This may also be a result of the broad model scale: perhaps models with a smaller spatial extent or more rich spatial structure would perform better in these areas, as they would not perform complete pooling of the estimated effects across a large heterogeneous region. Additionally, 2019 predictions are displayed interactively in **Supplement 1**. This type of interactive visualization could be used in disaster planning, with storm variable inputs parameterized by a forecast model to anticipate losses interactively in real time.

One limitation of this study is that the primary source of geospatial information is the beginning coordinates of the tornado and from these beginning coordinates, we performed weighted extractions to consider where a tornado may go. While this provides a predictive tool that requires less prior information (i.e. a specific path), an approach that explicitly considers tornado paths would be expected to provide better performance by providing a more accurate representation of areas impacted, and therefore a more high-fidelity representation of model inputs. Modeling this task as a sequence of input variables along a path which culminates in an observed property damage would be a highly interesting and potentially fruitful idea, perhaps using recurrent neural networks (Reichstein et al., 2019).

It is worth considering what it would take to operationalize this model for real-time use. In a real-time prediction task, it may be necessary to evaluate multiple future scenarios comprised of variable storm characteristics such as duration, spatial location, length, width, area, and the touchdown date-time at least. Constraining these scenarios or weighting different parameter combinations by a predictive forecast model might help to focus the predictions on relevant regions of parameter space. The operationalization of such a model is facilitated by the fact that all data used in this analysis are public, and all of the underlying software is open source. Just as tornado warnings are often acknowledged to help save lives (Hamillet al., 2005; Simmons and Stutter, 2005; Simmons and Stutter, 2008; Simmons and Sutter, 2012), accurate property damage predictions may help stakeholders avoid and better prepare for financial tragedies.

### 4.4. Conclusions

While there has been substantial research done in describing the characteristics of tornadoes and identifying variables of tornado risk, efforts to predict the economic damage that they cause or threaten to U.S. communities remains lacking. Here we show the potential that publicly available data sources and artificial neural networks have in such predictions, particularly that shallow but wide neural networks mimicking a traditional zero-inflated modeling approach provide promising results when utilizing storm, land cover, socioeconomic and demographic variables. When subsetted and evaluated on well-studied regions that allow comparison, these predictions are mostly consistent with existing literature on tornado impacts. In addition to two rounds of out-of-sample evaluation, we use these models to provide 2019 predictions via a prototype interactive dashboard to communicate the results to non-technical audiences.

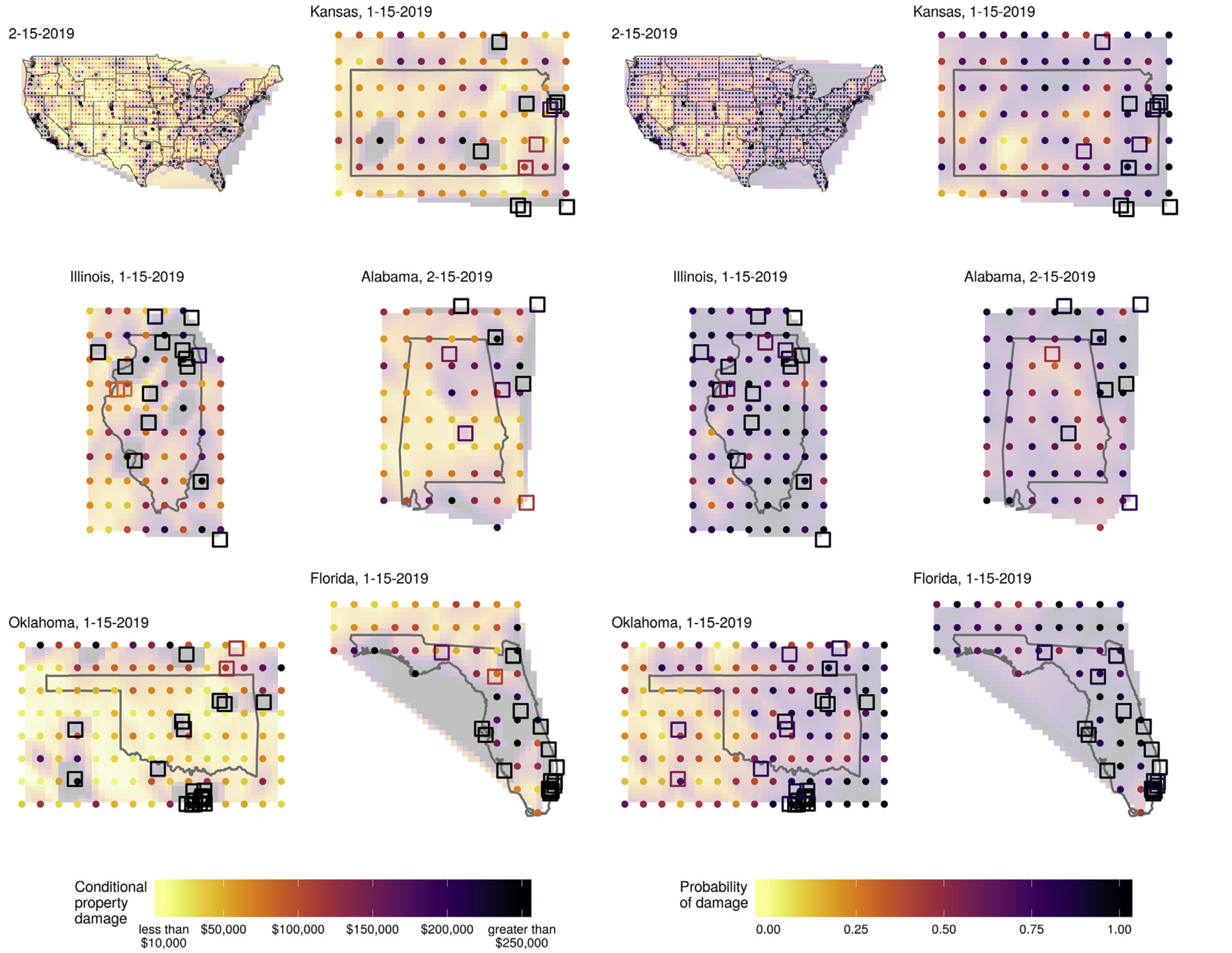

**Fig. 10.** Maps of 2019 predictions. The two left columns of maps represent predictions from the best conditional property damage model ($Y|Y > 0$), while the two right columns of maps represent the predictions from the best property damage occurrence models. Raw point data is plotted over a semi-transparent bilinear interpolation. Darker colors indicate larger predicted property damages and larger predicted probabilities of damage occurring, respectively. Titles indicate the state of focus (if applicable) and the date of displayed predictions. (For interpretation of the references to color in this figure legend, the reader is referred to the Web version of this article.)

## Acknowledgements

This research was made possible by the University of Colorado Boulder Grand Challenge initiative and the open source software community, with special thanks to the following open source software libraries: R (language - https://www.r-project.org/), Python (language - https://www.python.org/), tidyverse (R packages - https://www.tidyverse.org/), tidycensus (R package - https://walkerke.github.io/tidycensus/), SciPy (python packages - https://scipy.org/), rspatial (R packages - http://www.rspatial.org/), plot.ly (https://plot.ly/), and PyTorch (python framework - https://pytorch.org/).

## Appendix A. Supplementary data

Supplementary data to this article can be found online at https://doi.org/10.1016/j.wace.2019.100216.